%% file: paper.tex
\documentclass[sigconf]{acmart}

\usepackage[utf8]{inputenc}
\usepackage{xspace}
\usepackage{balance}  % for  \balance command ON LAST PAGE  (only there!)

\newcommand{\myparagraph}[1] { \noindent {\textbf {#1}} }

\newcommand{\prototype}{Visus}
\newcommand{\systemnamer}{\prototype\xspace}
\newcommand{\systemname}{\texttt{\prototype}\xspace}
\newcommand{\papertitle}{\systemnamer: An Interactive System for Automatic Machine Learning Model Building and Curation}

\renewcommand{\paragraph}[1]{\vspace{.08cm} \noindent \textbf{#1.}}
\newcommand{\paragraphtt}[1]{\vspace{.08cm} \noindent \textit{#1.}}

\def\withcolors{1}
\def\withnotes{1}

\ifnum\withcolors=1
  \usepackage{xcolor}
\fi

\ifnum\withnotes=1
  \usepackage[colorinlistoftodos,textsize=scriptsize]{todonotes}
\fi

\ifnum\withcolors=1
  \newcommand{\jfcolor}[1]{{\color{orange}#1}} % Juliana
  \newcommand{\ascolor}[1]{{\color{violet}#1}} % Aecio
\else
  \newcommand{\jfcolor}[1]{{#1}}
  \newcommand{\ascolor}[1]{{#1}}
\fi

\ifnum\withnotes=1
  \newcommand{\jfnote}[1]{\jfcolor{\textbf{JF: }\sf #1}}
  \newcommand{\asnote}[1]{\ascolor{\textbf{AS: }\sf #1}}
\else
  \newcommand{\jfnote}[1]{}
  \newcommand{\asnote}[1]{}
\fi
\newcommand{\ignore}[1]{\leavevmode\unskip} % eat unnecessary spaces before

\copyrightyear{2019}
\acmYear{2019}
\setcopyright{acmlicensed}
\acmConference[HILDA'19]{Workshop on Human-In-the-Loop Data
Analytics}{July 5, 2019}{Amsterdam, Netherlands}
\acmBooktitle{Workshop on Human-In-the-Loop Data Analytics (HILDA'19),
July 5, 2019, Amsterdam, Netherlands}
\acmPrice{15.00}
\acmDOI{10.1145/3328519.3329134}
\acmISBN{978-1-4503-6791-2/19/07}

\usepackage{balance}

\begin{document}
\title{\papertitle}

% Author list
\author{A\'ecio Santos}
\orcid{https://orcid.org/0000-0002-5124-7770}
\affiliation{
  \institution{New York University}
}
\email{aecio.santos@nyu.edu}

\author{Sonia Castelo}
\affiliation{
  \institution{New York University}
}
\email{s.castelo@nyu.edu}

\author{Cristian Felix}
\affiliation{
  \institution{New York University}
}
\email{cristian.felix@nyu.edu}

\author{Jorge Piazentin Ono}
\affiliation{
  \institution{New York University}
}
\email{jorgehpo@nyu.edu}

\author{Bowen Yu}
\affiliation{
  \institution{New York University}
}
\email{bowen.yu@nyu.edu}

\author{Sungsoo (Ray) Hong}
\affiliation{
  \institution{New York University}
}
\email{rayhong@nyu.edu}

\author{Cl\'audio T. Silva}
\affiliation{
  \institution{New York University}
}
\email{csilva@nyu.edu}

\author{Enrico Bertini}
\affiliation{
  \institution{New York University}
}
\email{enrico.bertini@nyu.edu}

\author{Juliana Freire}
\orcid{https://orcid.org/0000-0003-3915-7075}
\email{juliana.freire@nyu.edu}

\affiliation{
  \institution{New York University}
}

\begin{abstract}
  \input{sections/00-abstract}
\end{abstract}

%
% ACM CCS categories: https://dl.acm.org/ccs/ccs.cfm
%



\maketitle

\input{sections/01-introduction}

\input{sections/02-relatedwork}

\input{sections/03-framework}

\input{sections/04-system}

\input{sections/05-scenarios}

\input{sections/06-userfeedback}

\input{sections/98-conclusion}

\input{sections/99-acknowledgments}

\balance

\bibliographystyle{ACM-Reference-Format}
\bibliography{paper}

\end{document}

%% file: sections/00-abstract.tex
While the demand for machine learning (ML) applications is booming, there is a scarcity of data scientists capable of building such models.
Automatic machine learning (AutoML) approaches have been proposed that help with this problem by synthesizing end-to-end ML data processing pipelines.
However, these follow a best-effort approach and a user in the loop is necessary to curate and refine the derived pipelines.
Since domain experts often have little or no expertise in machine learning, easy-to-use interactive interfaces that guide them throughout the model building process are necessary.
In this paper, we present \systemname, a system designed to support the model building process and curation of ML data processing pipelines generated by AutoML systems.
We describe the framework used to ground our design choices and a usage scenario enabled by \systemname. Finally, we discuss the feedback received in user testing sessions with domain experts.

%% file: sections/01-introduction.tex
\section{Introduction}
\label{sec:intro}

% Intro: Machine learning is booming
Machine Learning (ML) models are increasingly being adopted
in many applications.
%, including traffic and weather prediction,
% news and product recommendation, spam filtering, and fraud detection,
% to name a few.
% Problem: ML demand grows larger than data scientists supply
However the supply of data scientists capable of
building such models has not kept up with the demand~\cite{forbes-data-science-demand}.
% Current solutions: AutoML
Recent approaches to automatic machine learning (AutoML) help address this problem by automating not only model fitting, but also the synthesis of end-to-end ML data processing pipelines, from data loading, data cleaning, and feature engineering to model fitting and selection, and hyper-parameter tuning \cite{olson@gecco2016, drori@2018, gil@automl2018}.
Commercial solutions such as Google's Cloud AutoML \cite{google-cloudml} and Microsoft's Azure Machine Learning Studio \cite{ms-azure-ml-studio} are also available.
AutoML systems require as input only the training data and a well-defined problem specification that describes the target variables to be predicted, the data attributes to be used as features, and the performance metric to be optimized.
By automating  many of the steps of the ML model building process, AutoML systems allow data scientists to solve problems more efficiently.

% Motivation: the democratization of ML to wider audiences
However, AutoML systems are not sufficient to democratize ML to wider audiences.
To empower domain experts, who have little or no expertise in ML, to develop ML pipelines easy-to-use interactive interfaces are needed  that guide them throughout the model building process~\cite{cashman2018, gil@iui2019}. 
Additionally, AutoML systems put their best effort to generate pipelines that optimize a given performance metric, which often result in black-box models that cannot easily be understood by humans.
This, coupled with the ever-increasing concerns about the societal impact of data-driven decision making~\cite{ zliobaite@arxiv2015, zafar@www2017, hardt@nips2016}, underscores the importance of bringing the human in the loop of the model building process.
Enabling domain-experts to build ML models not only has the potential to lessen the demand for data scientists, but it also enables experts that possess domain knowledge to assess, validate, and potentially improve model outcomes.

% Challenges and limitations
Prior research in the field of visual analytics --- which has been regarded as a grand challenge \cite{thomas@book2005} --- has proposed multiple frameworks and systems that integrate ML models and interactive visual interfaces to facilitate analytical reasoning~\cite{cashman2018}.
However, these systems typically have to integrate tightly with the model they use to provide insights about the model behavior.
With the inclusion of an AutoML system, we cannot assume any prior knowledge about inner-workings of the returned models because of the large variety of components present in the derived ML pipelines.

% Our solution
In this work, we propose a new framework to support the model building process and curation of automatically generated predictive data processing pipelines.
Similar to prior work~\cite{cashman2018}, our framework is iterative and includes multiple components that support exploratory data analysis (EDA), problem specification, model generation and selection, and confirmatory data analysis (CDA).
Additionally, we propose new components that allow users to 1) search the Web for new relevant datasets to augment the input data, and 2) summarize, compare, and explain models generated by AutoML systems.  
The first component enhances the original data by adding new records (through union operations) or new attributes (through joins) that can potentially improve the model performance.  The second component  allows users to better understand the automatically generated pipelines and to select the one that best fulfills their intent.  Finally, we present \systemname, a system that implements our proposed framework and integrates visual analytics techniques in the model building and curation process. We also describe a usage scenario that illustrates the use of  \systemname to support data analysis and decision-making processes and discuss the feedback received in user testing sessions.

Our contributions can be summarized as follows:
\begin{itemize}
  \item We propose a new visual analytics framework to support the process of building and curating ML pipelines automatically generated by AutoML systems;
  \item We propose new components for 1) searching datasets on the Web to augment the original dataset, and 2) support model selection through model summarization, comparison, and explanation;
  \item We present \systemname, an interactive system that implements our proposed framework. We describe usage scenarios enabled by \systemname that show how a domain-expert can build a regression model to support decision-making.
\end{itemize}

%
% Organization of the paper
%
\paragraph{Outline} The remainder of this paper is structured as
follows.  We discuss related work in Section~\ref{sec:related}.
In Section~\ref{sec:framework}, we describe our proposed visual analytics framework.
In Section~\ref{sec:system}, we describe \systemname and our design choices. %how it enables domain experts to build ML models.
We present a usage scenario in Section~\ref{sec:scenarios}, discuss the user feedback in Section~\ref{sec:user-feedback}, and conclude in Section~\ref{sec:conclusion}, where we outline directions for future work.

%% file: sections/02-relatedwork.tex
\section{Related Work}
\label{sec:related}

This paper explores a problem that lies at the intersection of Automatic Machine Learning (AutoML) and Human Guided Machine Learning (HGML). In what follows, we briefly describe the state of the art in these two fields and how they can be connected to achieve our goal. For a comprehensive literature review, see \citet{hutter2019automatic} and \citet{gil@iui2019}. 

Creating machine  learning pipelines is a time consuming task that demands domain expertise and machine learning knowledge. These pipelines usually contain many steps, including data pre-processing, algorithm selection and hyper-parameter tuning. AutoML aims to automate the design and implementation of ML pipelines, reducing the time and level of expertise necessary to deploy solutions for real~world problems \cite{hutter2019automatic}. Some examples of state of the art AutoML methods include Autosklearn~\cite{feurer2015efficient},  TPOT~\cite{olson@gecco2016}, and AlphaD3M ~\cite{drori@2018}. Autosklearn and TPOT are able to automatically generate pipelines for classification, and AlphaD3M can tackle both classification and regression problems. As for the pipeline search strategies, Autosklearn uses Bayesian optimization combined with meta-learning, TPOT uses genetic programming and tree-based pipelines, and AlphaD3M uses reinforcement learning together with Monte-Carlo Tree Search and self play.

While AutoML speeds up the pipeline generation process and helps to democratize machine learning, humans still have an important role to play: by interacting with the AutoML system, humans can provide  additional knowledge to the algorithm and constrain the space of possible models so that they are more suitable for a specific problem. For example, experts can specify important features or pre-processing steps that need to be applied in the data~\cite{hutter2019automatic, gil@iui2019}. To this end, HGML systems combine  AutoML with specialized Visual Analytics (VA) systems, enabling humans to interact with the pipeline search and improve the model prediction results. \citet{gil@iui2019} presented a task analysis and desiderata for HGML and assessed how current AutoML and VA systems attempt to solve these tasks. The  authors grouped desired features for HGML systems under three major categories: \emph{data use}, specifying how features and samples should be used, and augmenting the data with new features, \emph{model development}, constraining the search space of possible models, and \emph{model interpretation}, interpretation, diagnosis and comparison of machine learning models. Implementing the HGML desired features individually is possible, and described by the authors. However, implementing a VA system that combines all requirements remain a challenge, given the iterative and nonlinear nature of the interactions. 

Two recent implementations of HGML systems are \emph{Snowcat}~\cite{cashman2018} and \emph{Two Ravens}~\cite{honaker2014statistical, gil@iui2019}. Both systems enable data exploration  and visualization. Furthermore, they allow the basic data use specification, such as setting which features to use to solve a problem, and a rudimentary comparison of the generated ML models  using traditional metrics and visualizations, such as accuracy and confusion matrix. Comparing the capabilities of these systems with the desiderata proposed by \citet{gil@iui2019}, we note two major gaps. First, neither system supports the augmentation of data, a feature that can increase the amount of information the models have access to and potentially improve the performance of the generated pipelines.  Second, they lack  support for model interpretation beyond simple ML metrics. Our system, \systemname, aims to fill this gap in the current HGML systems. In the next session, we describe how we integrated data search and automatic joins in our framework \cite{chapman@arxiv2019}, as well as model interpretation based on rule visualization \cite{ming@tvcg2019}.

%% file: sections/03-framework.tex
\section{Proposed Framework}
\label{sec:framework}

\begin{figure}[t]
  \centering
  \includegraphics[width=0.48\textwidth]{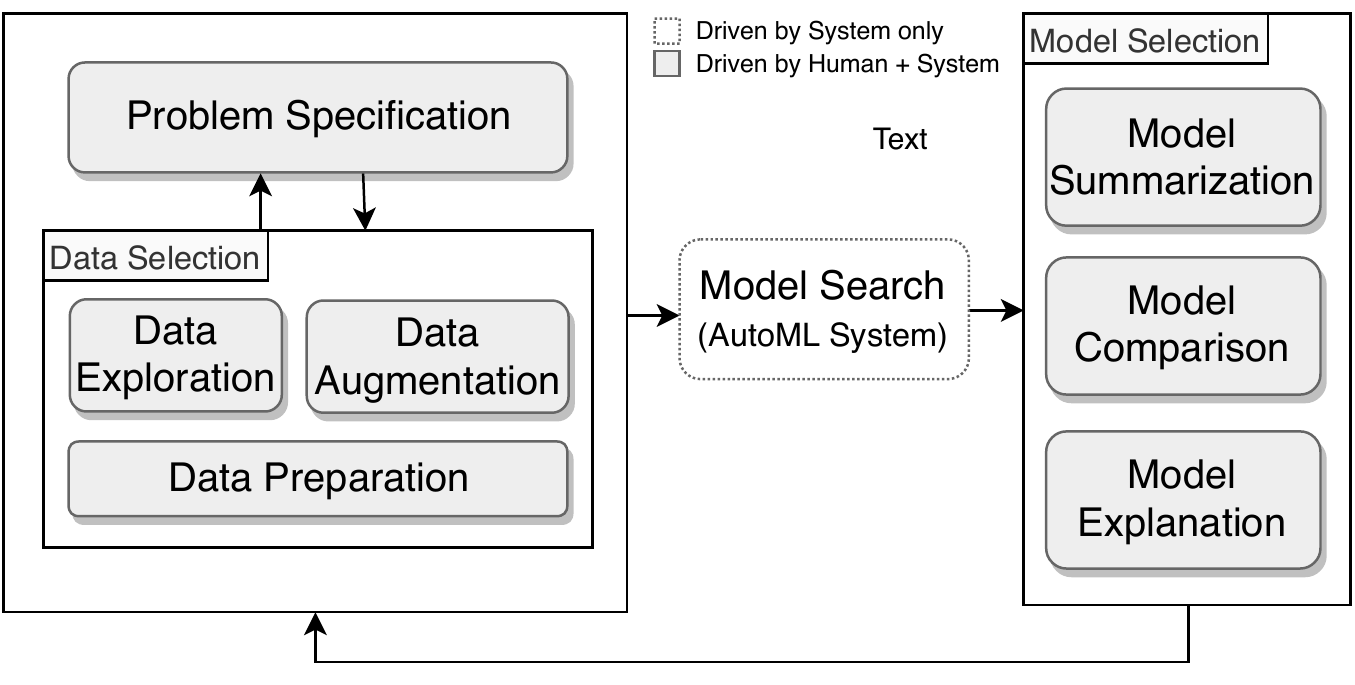}
  \caption{Our proposed framework encompasses problem definition, model generation, and model selection.}
  \label{fig:framework}
  \vspace{-0.5cm}
\end{figure}

%In this section, we describe our proposed visual analytics framework.
Figure~\ref{fig:framework} shows the main components of our visual analytics framework.
Each rounded box denotes a task and the connecting arrows denote the user workflow.
In what follows, we describe each of the components in detail.

% problem definition?
\paragraph{Problem Specification}
The first challenge we address in our framework is the translation of a problem in the user domain (e.g., \textit{What are the important factors that lead to traffic accidents?}, or \textit{What genes or environmental conditions influence a plant's growth?}) to a machine-readable problem specification that can be fed to an AutoML system.
A problem specification consists of a set of well-defined properties that describe the users' analysis goal: a problem type (e.g., classification, regression, clustering, time series forecasting, etc.), a target attribute to be predicted (e.g., a categorical attribute containing classes, a numerical attribute with values), a performance metric to be optimized (e.g., precision, recall, F$_1$, mean squared error,  etc.), and a list of data attributes to be used as features.
Additionally, it may contain other parameters such as time constraints (i.e., for how long the pipeline search should run), the maximum number of pipelines to be generated, the performance metrics to be reported and the evaluation method for computing them (e.g., $k$-fold cross-validation, hold-out test set).

% data exploration and transformation
\paragraph{Data Selection}
In order to create a problem specification, the user first needs to select the data required
to solve the problem.
There are three data-related components that support the problem specification task: \textit{data exploration}, \textit{data preparation}, and \textit{data augmentation}.
We arrange them in a group, namely ``Data Selection'', because they are intertwined, in the sense that not only they support the problem specification task, but also are influenced by problem specification.
For example, consider the \textit{data exploration and preparation} components. In order to specify a problem, a user might need to first perform an exploratory data analysis to better understand the data.
During the course of exploration, the user may realize that an attribute contains outliers which need to be removed, or that an attribute contains missing data that should be filled out with a default value.
The user may also decide to update the problem specification to indicate that some attributes should be excluded from the model.
% data augmentation

The \textit{data augmentation} component allows the user to search for relevant datasets that could potentially supplement the original data with new records or attributes and, thus, improve the model performance.
Describing in details the inner-workings of the dataset search engine we use to support augmentation is out of the scope of this paper.  However, it is important to note that it is critical that the search results returned by the dataset search engine should not only be relevant to the domain of interest, but also should be appropriate for integration (via join or union operations) with the original dataset.
The main role of the visual interface is then to allow the user to interact with the search results and facilitate dataset relevance prediction (i.e., the search interface should display effective dataset summaries that allow for the user to decide whether the dataset discovered is relevant and should be merged to the original dataset).

\paragraph{Model Search}
After the data is selected and the problem is specified, the AutoML system can be invoked to carry out the pipeline search phase. Note that this is the only task in our framework which there is no human engagement.

\paragraph{Model Selection}
AutoML systems aim to generate a model that optimizes a given performance metric in the training dataset. 
However, the best-scoring model learned using the training set is not always the best model.
When the training examples are not diverse and are not good representatives of the real sample distribution, the learned models may not generalize well for samples observed in real-world, and thus the performance might drop in a held-out validation set.
Additionally, some evaluation metrics may have trade-offs.
For instance, consider the pipelines generated for a binary classification problem. The best pipeline in terms of precision may have a low recall score.
Instead of blindly selecting only the best performing pipeline --- and given that often a large number of models is evaluated during the optimization process ---  a better approach for model selection could be to collect a sample of the best-performing pipelines and then present them to a domain-expert to pick the best one.
%
% JF: Unlike traditional visual analytics approaches to what? 
Unlike traditional visual analytics approaches, a system that includes an AutoML component must be able to handle a large number of ML pipelines.
Therefore, in order to support model selection, the user interface should provide visualizations to facilitate exploration, comparison, and explanation of multiple models.
Our framework includes three components to assist in this task as detailed next.

\paragraphtt{Model summarization}
Since AutoML systems create many pipelines, the end goal of model summarization should be to facilitate the selection of a smaller set of pipelines for further inspection. Visualizations should aim to provide concise representations of a set of pipelines and the ability to easily select the best-performing ones. 
% For instance, they could 1) provide an overview of generated pipelines and their performances, and 2) make evident performance metric trade-offs when they exist. 

\paragraphtt{Model comparison}
Once a sample of the best-performing pipelines is selected, the user is able to compare them in detail.
Visualizations should highlight the commonalities and differences between two or more pipelines. More specifically, pipelines may differ not only in performance metric scores but also in their data processing steps, and how they make correct or wrong predictions.

\paragraphtt{Model explanation}
The user might also want to gain a better insight into how models make predictions and how the prediction accuracy is distributed conditioned to specific attributes or the target variable.
%this category of pipelines includes visualizations that aim to gain a better insight on how models make predictions.
For instance, one might be interested in understanding what is the impact of certain feature combinations to a prediction outcome. Therefore, explanations might help not only to validate whether the model is making correct predictions for the right reasons but also to learn about data relationships that were not known a priori.

%% file: sections/04-system.tex
\section{The \systemnamer System}
\label{sec:system}

With \systemname, we aim to aid domain experts to interactively formulate prediction problems, build models, and select a suitable model among a set of automatically generated models.
\systemname provides an easy-to-use interface where users can 1) define a problem, 2) explore summaries of their input dataset, 3) perform data augmentation, 4) explore and compare models with respect to their performance scores and prediction outputs.

\begin{figure*}[t]
    \centering
    \includegraphics[width=0.85\textwidth]{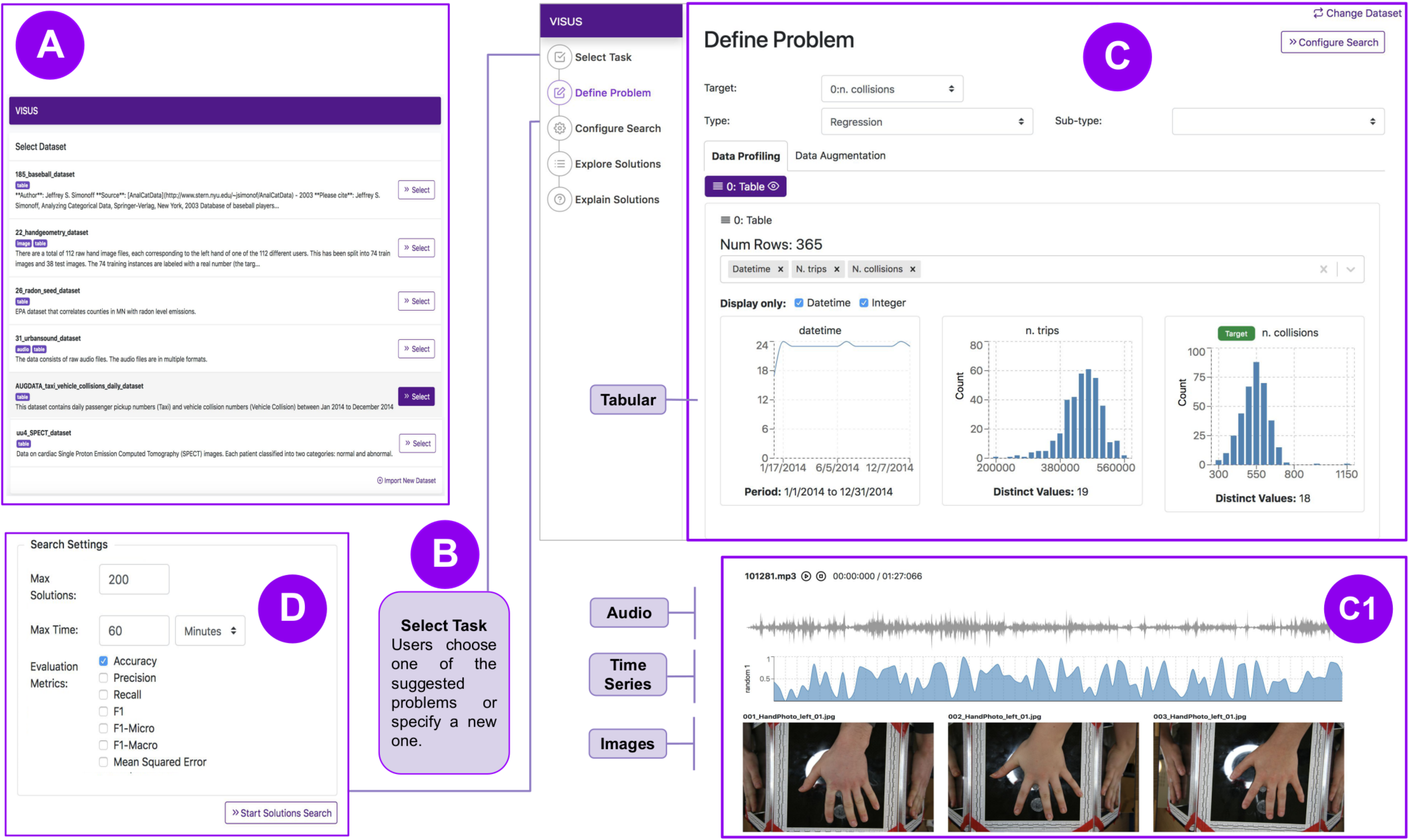}
    \caption{\systemnamer's data selection and problem definition screens:  (A) select or load dataset view, (B) select task view (create a new problem, or load an previously created problem), (C) define problem view (displays data summaries and allows selection of problem parameters), (D) configure search view (allows setup of additional model search parameters).}
    \label{fig:defineproblem}
\end{figure*}

\paragraph{Define Problem and Explore Dataset}
%Here we describe the system features related to the problem specification and data selection tasks from our framework.
As Figure~\ref{fig:defineproblem} illustrates, the operations for problem specification and data selection  are implemented in four screens (views) in \systemname: \textit{Select Dataset}, \textit{Select Task}, \textit{Define Problem}, and \textit{Configure Search}.
% We describe in detail each of these screens in what follows.

\paragraphtt{Select Dataset} 
Allows the user to select or load a dataset (Figure~\ref{fig:defineproblem}(A)).

\paragraphtt{Select Task}
Users can either create a new problem specification or select one of the existing problems specifications that have been created previously (Figure~\ref{fig:defineproblem}(B)).

\paragraphtt{Define Problem}
To construct good predictive models, it is important to identify data attributes (features) that lead to accurate predictions.
For this, \systemname profiles the data and displays visualizations that summarize the information, helping users to obtain insights about the features.
\systemname supports multiple types of datasets, including tabular data, time series, audio, and images (see Figure~\ref{fig:defineproblem}(C1)).
For all of them, \systemname provides a \textit{select input control} for adding/removing data on demand with auto-complete and multi-select functionalities to help users to easily find data attributes or files.

For tabular data, \systemname generates visualizations that adapt to different attribute data types.  For instance, for numerical and categorical attributes, \systemname generates descriptive statistics and summarizes data values using a histogram.  When visualizing time series datasets, \systemname provides a stacked time series plots along with an interactive \textit{brushing} tool to help users to explore and visualize different time ranges.  For audio, \systemname lets users reproduce the audio and display its waveform, and for images, it uses thumbnails to show a sample of the images present in the dataset.

\paragraphtt{Configure Search}
In this screen, users set up additional parameters of the model search such as a maximum number of solutions (pipelines) to be generated by the AutoML system, a maximum wait time to conclude the model search.
Furthermore, users can also specify the performance metrics they want to use for evaluating the solutions generated.

\begin{figure}[t]
    \centering
    \includegraphics[width=0.48\textwidth]{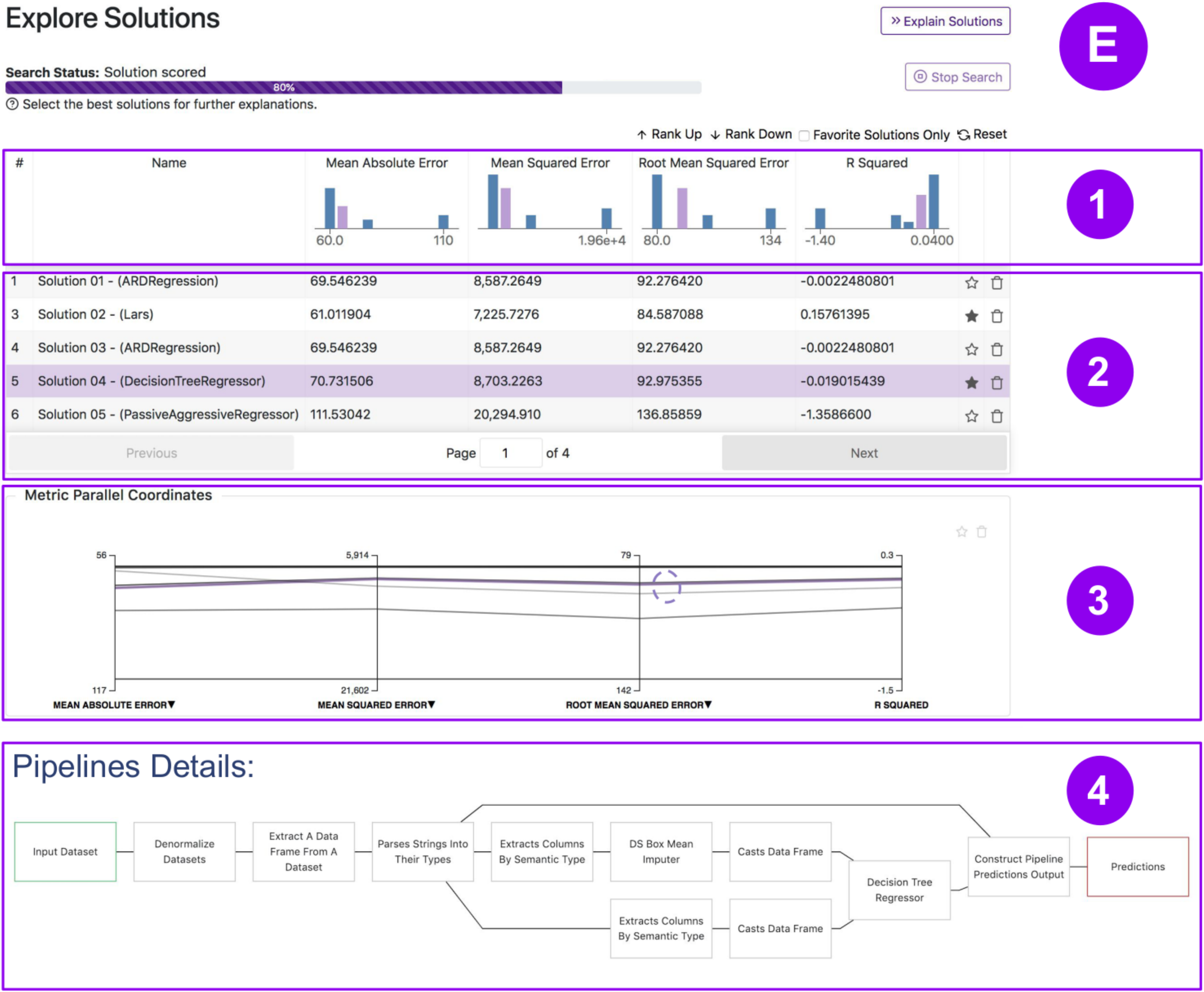}
    \caption{Explore models view}
        \label{fig:explore-solutions}
        \vspace{-0.5cm}
\end{figure}

\paragraph{Inspect Model Summaries}
To allow exploration of the pipelines generated by the Model Search, \systemname displays them in a solution table as shown in Figure \ref{fig:explore-solutions}(E2).
In this table, users can see the different solutions and their associated performance metrics.
Users can also sort solutions by metric, allowing them to quickly identify the best solutions according to each metric.
Additionally, \systemname provides a histogram of the scores associated with each performance metric, which allows the users to visualize the distribution of scores of all generated solutions (see Figure \ref{fig:explore-solutions}(E1)).
When the user hovers over a solution in the table, the bin associated with the score of the hovered solution changes its color, allowing quick comparison of its performance against all other solutions.
Furthermore, \systemname provides a parallel coordinates chart, as shown in Figure \ref{fig:explore-solutions}(E3). Each vertical line corresponds to a performance metric and each horizontal line corresponds to a pipeline. Such a visualization allows for quick identification of metric trade-offs. If the user hovers over a pipeline, he/she can see how its performance measures compare to those of other pipelines. Finally, users can also inspect the actual data processing steps of each pipeline, including their learning algorithms and pre-processing primitives  (see Figure~\ref{fig:explore-solutions}(E4)). 

\begin{figure}[t]
	\centering
	\includegraphics[width=0.48\textwidth]{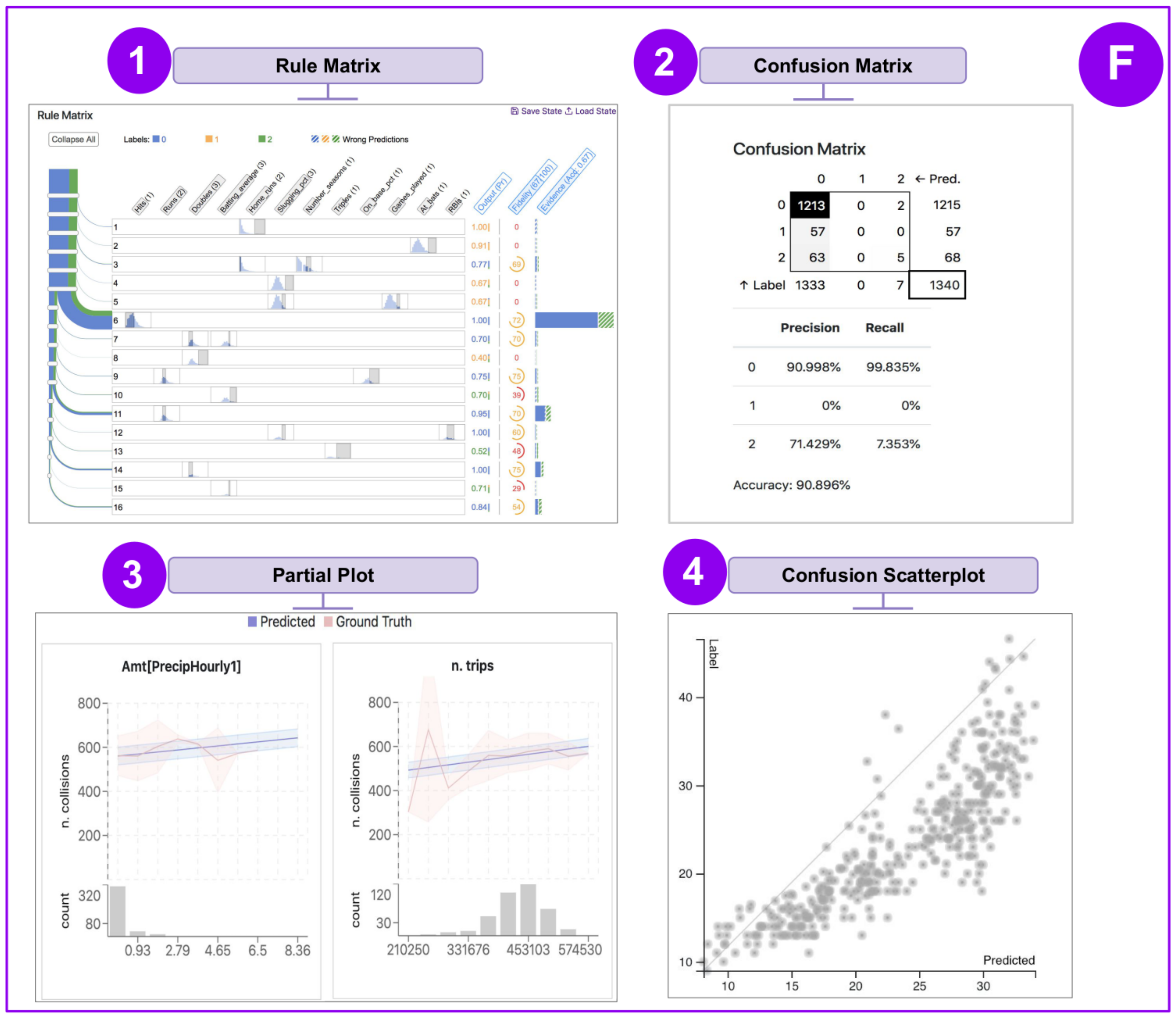}
	\caption{Model explanation support for classification models: (1) rule matrix plot and (2) confusion matrix, and for regression models: (3) partial dependence plot and (4) confusion scatter plot.}
	\label{fig:explanations}
	\vspace{-0.1cm}
\end{figure}

\paragraph{Examine Model Explanations}
To better understand how a given pipeline performs, \systemname generates more detailed explanations.
For classification problems, \systemname currently supports two visualizations: a standard confusion matrix and rule matrix~\cite{ming@tvcg2019}.
The confusion matrix shows the predicted classes as columns and the true classes as rows as shown in Figure \ref{fig:explanations}(F2).
The rule matrix uses a surrogate model to learn classification rules that explain the behavior of the model.
It displays rules as rows and features as columns (see Figure~\ref{fig:explanations}(F1)).
To explain regression problems, \systemname supports two visualizations: a partial dependence plot \cite{hastie2009elements} and a confusion scatter plot.
The partial dependence plots show on the bottom the distribution of samples for each attribute value range, and on the top the predictions and ground truths for these values (see Figure \ref{fig:explanations}(F3)).
The confusion scatter plot encodes each data sample as a dot, the predictions in the $y$-axis, and the value in the $x$-axis. Figure \ref{fig:explanations}(F4) shows an example.

\begin{figure}[t]
	\centering
	\includegraphics[width=0.48\textwidth]{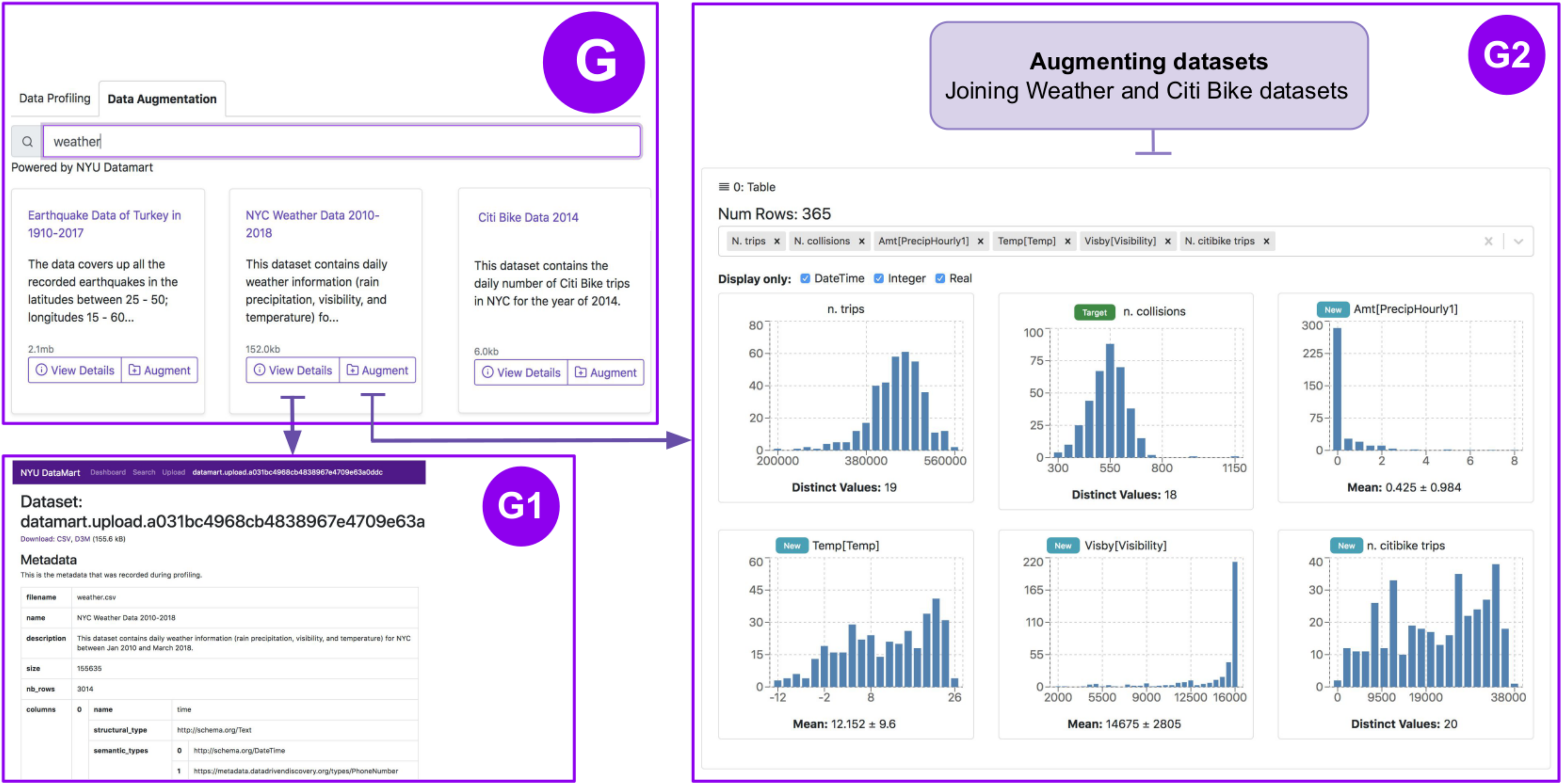}
	\caption{Data augmentation support in \systemnamer. (G) Dataset search results page, (G1) dataset details, and (G2) define problem screen showing data attribute summaries after data augmentation.}
	\label{fig:data-augmentation}
	\vspace{-0.5cm}
\end{figure}

\paragraph{Data Augmentation}
In the \textit{Define Problem} view (see Figure \ref{fig:defineproblem}(C)), the user can navigate to the \textit{data augmentation} tab, where he/she can search for related datasets that could be joined with (join operation) or appended (union operation) to the original data (see Figure~\ref{fig:data-augmentation}(G)). Before performing augmentation, users can view details about the suggested datasets by pressing the \textit{View Details} button, as shown in Figure~\ref{fig:data-augmentation}(G1).

%% file: sections/05-scenarios.tex
\section{Use Case}
\label{sec:scenarios}

In this section, we describe a data analysis scenario that \systemname enables. 
Consider the following example. As part of the New York City Vision Zero Initiative \cite{nyc-vision-zero}, an expert from the Department of Transportation is trying to devise policies to reduce the number of traffic fatalities and improve safety in NYC streets.

\begin{figure}[t]
	\centering
	\includegraphics[width=0.45\textwidth]{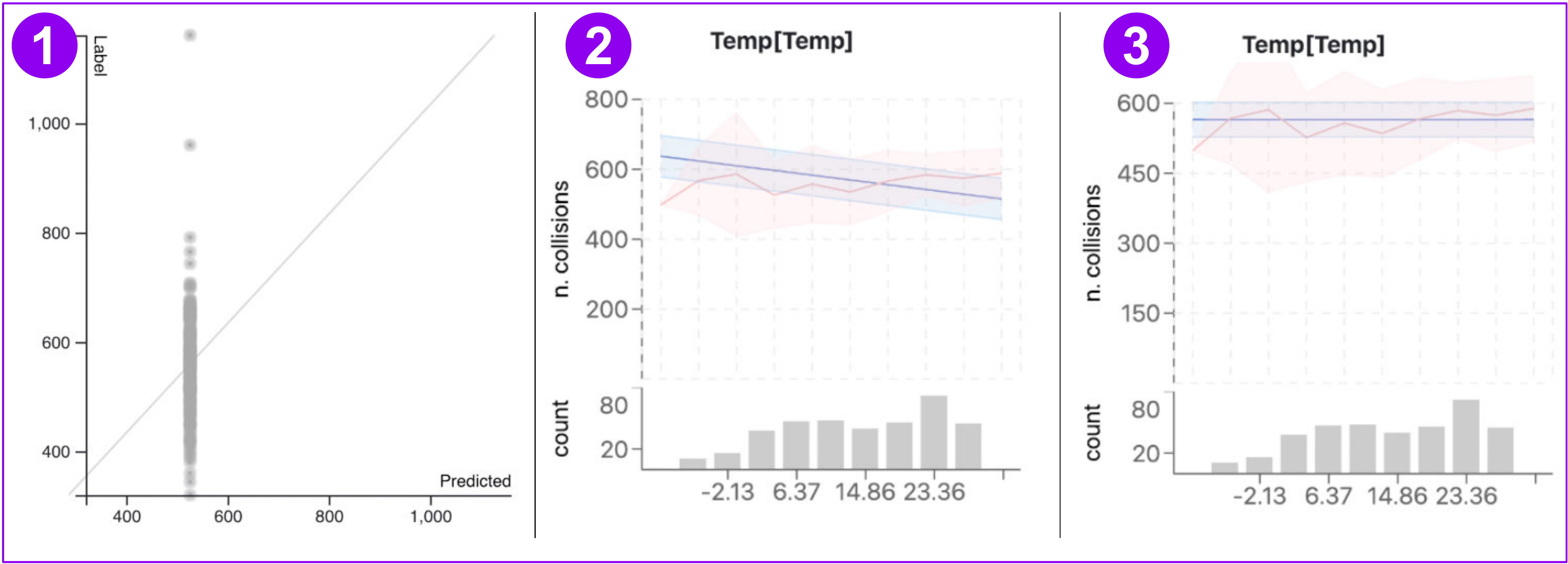}
	\caption{User-generated visualizations: (1) initial confusion scatter plot, (2) partial dependency plot for temperature from model A, (3) partial dependency plot for temperature from model B.}
  \label{fig:scenarios}
 	\vspace{-0.5cm}
\end{figure}

\paragraph{Data exploration and model search specification}
Initially, the expert loads a dataset about traffic collisions and taxi trips.
Her goal is to create a model that predicts the number of collisions; such a model can allow her to explore what-if scenarios.
After loading her dataset, \systemname profiles it and displays visualizations that summarize the data.
She will see a screen similar to Figure~\ref{fig:defineproblem}(C)  which displays the different features, namely the \textit{date}, \textit{number of trips}, and \textit{the number of collisions}, and how their values are distributed.
In this same screen, she can also select the \textit{target} variable she wants to predict and the type of model she wants to build (e.g., classification, regression).
After pressing the ``Configure Search'' button, she is presented another screen where she can specify additional parameters of the model search specification (e.g., the maximum running time, evaluation metrics).
The next step ("Start Solution Search") triggers the execution of the AutoML system and a new  screen is shown to display the results (``Explore Solutions'').
As soon as models are generated by the AutoML system, \systemname displays them in the ``Solutions Table'', shown in Figure~\ref{fig:explore-solutions}, 
which displays the generated pipelines and summarizes their evaluation scores in a histogram.
To get more insight into the behavior of the different solutions, she moves to the next step,  ``Explain Solutions'', which produces explanatory visualizations such as the confusion scatter plot as seen in Figure~\ref{fig:scenarios}(1).
Analyzing the confusion scatter plot she notices that, for the pipeline with the smallest error, all instances are being predicted with the same values (all are in the same position on the $x$-axis).
Clearly, this model is not very useful. Thus, she takes a step back and returns to the ``Define Problem'' screen to refine the problem.

\paragraph{Augmenting the data to increase performance}
Now she hypothesizes that the unsatisfactory performance may be a result of insufficient predictive power of her data attributes.
She then clicks on the tab ``Data Augmentation'', to search for additional data.
Intuitively, weather information, such as bad weather conditions can potentially increase the number of collisions.
She uses the search form to find datasets that have the keyword ``weather'' and that are related to her input data, i.e., that can be joined with it,  and then selects a matching dataset.
Because many collisions also involve cyclists, she also searches for datasets related to her input data using the keyword ``\texttt{citi bike}'', hoping to find a dataset containing information about the number of cyclists in the streets.
Because both weather and citi bike datasets contain temporal attributes, as has her original data, \systemname is able to automatically perform a temporal join of these datasets.
After doing the augmentations with weather and Citi Bike data, her data increased from 3 to 7 features.
She now re-runs the solution search to generate solutions using the augmented data and finds that the best pipeline has now improved from 71.29 to 61 in terms of mean absolute error, a significant improvement.

After picking her new favorite pipelines on the ``Explore Solution'' screen, she inspects them in more detail.  She notices that now her best pipeline's predictions improved and the confusion scatter plot shows a more diagonal pattern (see Figure \ref{fig:explanations}(F4)), which is the desired behavior.  By using the partial dependency plot, she inspects the impact of the features in different models. She notices that, for pipeline A, the \textit{temperature} has a strong negative correlation with the number of collisions (Figure~\ref{fig:scenarios}(2)), while the pipeline B seems not to use temperature, as indicated by a flat line (Figure~\ref{fig:scenarios}(3)) showing no correlation.  Both pipelines use different versions of the Lasso algorithm and have similar performance in terms of error, she uses her domain knowledge, and decides to choose the model B which does not take temperature into account, even though its performance according to evaluation metrics is inferior when compared to A.

%% file: sections/06-userfeedback.tex
\section{User Feedback}
\label{sec:user-feedback}

All our design decisions discussed previously have been grounded on feedback from domain-experts --- with little or no data science knowledge --- who interacted with earlier versions of our system. In this section, we discuss the feedback received from users in more recent sessions. All users interacted with the system using data that comes from their area of expertise and that they were already familiar with. 
For instance, one of the users has a Ph.D. in Biology who is interested in plant genetics and phenotyping and another is a manager at a U.S. agency who works with forward operators, soldiers, and analysts who want to analyze their data.

Users were generally satisfied with the system design, and were keen to start making connections to additional augmented data, wanted to overlay weather, add genetic information, and even geolocation.
However, we noticed that there is still a gap in the communication of machine learning concepts to domain experts.
For instance, it may not be clear for the users the meaning of evaluation metrics. \textit{What is the difference between $R^2$,
mean absolute error, and mean squared error? When should one choose one over the other?}

Users also had some difficulty to understand and read the classification model explanations.
For example, visual explanations like the Rule Matrix \cite{ming@tvcg2019} summarize complex Bayesian classification rules derived from a surrogate model in a single visualization.
In order to understand them properly, details had to be explained, including the meaning of a rule and statistical concepts such as fidelity and evidence.
Although users claimed to understand the visualization after verbal explanations from an expert, it was not intuitive for them at first glance.
This suggests that while such visualizations may be suitable for data scientists with machine learning knowledge, further research is necessary towards visual explanations geared towards non-expert audiences.

%% file: sections/98-conclusion.tex
\section{Conclusion}
\label{sec:conclusion}

In this paper, we presented \systemname, an interactive system that allows users to build and curate end-to-end machine learning pipelines.  We described the framework and design choices that ground our step-by-step interface design and system workflow.  We also introduced new components that allow users to perform interactive data augmentation and visual model selection.  Finally, we conducted user testing sessions with domain experts with little machine learning expertise to evaluate our system.  The results suggest that while users were satisfied with the overall system design, more research is necessary towards bridging the communication gap between machine learning concepts and users with little knowledge of machine learning.  As future work, we plan to explore new, intuitive visual explanations that are able to explain how ML models make predictions to wider audiences.  Additionally, there are many open questions that need to be investigated to design interfaces for data augmentation that support users in deciding which datasets are relevant for their problem.

%% file: sections/99-acknowledgments.tex
\vspace{10pt}
\myparagraph{Acknowledgments.}
This work was partially supported by the DARPA D3M program,
and NSF award CNS-1229185.
Any opinions, findings, and conclusions or recommendations expressed in
this material are those of the authors and do not necessarily reflect
the views of NSF and DARPA.